\documentclass[runningheads]{llncs}
\usepackage{graphicx}
\usepackage{comment}
\usepackage{amsmath,amssymb} 
\usepackage{color}

\usepackage{color, colortbl}
\usepackage{array}
\definecolor{Gray}{gray}{0.8}
\def\x{{\mathbf x}}

\usepackage{graphics} 
\usepackage{epsfig} 
\usepackage{amsmath} 
\usepackage{amssymb}  
\usepackage{multirow}
\usepackage{comment}
\usepackage{lipsum}

\usepackage{wrapfig}
\usepackage{floatrow}

\def\x{{\bf x}}
\def\y{{\bf y}}

\def\Re{\mathbb{R}}

\newcommand{\m}[1]{\ensuremath{\mathcal{#1}}}
\newcommand{\h}[1]{\tilde{#1}}
\newcommand{\bbold}[1]{{[\bf #1]}}

\begin{document}
\pagestyle{headings}
\mainmatter
\def\ACCV20SubNumber{539}
\title{Semi-supervised Facial Action Unit Intensity Estimation with Contrastive Learning}

\titlerunning{Semi-supervised AU Intensity Estimation with Contrastive Learning}
%
\author{Enrique Sanchez \and
Adrian Bulat \and
Anestis Zaganidis \and
Georgios Tzimiropoulos}
\authorrunning{E. Sanchez et al.}
%
\institute{Samsung AI Center, Cambridge, UK \\
\email{\{e.lozano, adrian.bulat, a.zaganidis, georgios.t\}@samsung.com}}
\maketitle

\begin{abstract}
This paper tackles the challenging problem of estimating the intensity of Facial Action Units with few labeled images. Contrary to previous works, our method does not require to manually select key frames, and produces state-of-the-art results with as little as $2\%$ of annotated frames, which are \textit{randomly chosen}. To this end, we propose a semi-supervised learning approach where a spatio-temporal model combining a feature extractor and a temporal module are learned in two stages. The first stage uses datasets of unlabeled videos to learn a strong spatio-temporal representation of facial behavior dynamics based on contrastive learning. To our knowledge we are the first to build upon this framework for modeling facial behavior in an unsupervised manner. The second stage uses another dataset of randomly chosen labeled frames to train a regressor on top of our spatio-temporal model for estimating the AU intensity. We show that although backpropagation through time is applied only with respect to the output of the network for extremely sparse and randomly chosen labeled frames, our model can be effectively trained to estimate AU intensity accurately, thanks to the unsupervised pre-training of the first stage. We experimentally validate that our method outperforms existing methods when working with as little as $2\%$ of randomly chosen data for both DISFA and BP4D datasets, without a careful choice of labeled frames, a time-consuming task still required in previous approaches. 
\end{abstract}

\section{Introduction}
\label{sec:intro}
\noindent Facial actions are one of the most important means of non-verbal communication, and thus their automatic analysis plays a crucial role in making machines understand human behavior. The set of facial actions and their role in conveying emotions has been defined by the Facial Action Coding System (FACS~\cite{ekman02}). FACS define a set of atomic facial movements, known as Action Units, whose combination can be correlated with both basic and complex emotions. Action Units are categorically modeled according to their intensities, with values that range from $0$, indicating the absence of an AU, to $5$, indicating the maximum level of expressivity of an AU. 

 \begin{figure*}[t!]
    \centering\includegraphics[width=1\linewidth]{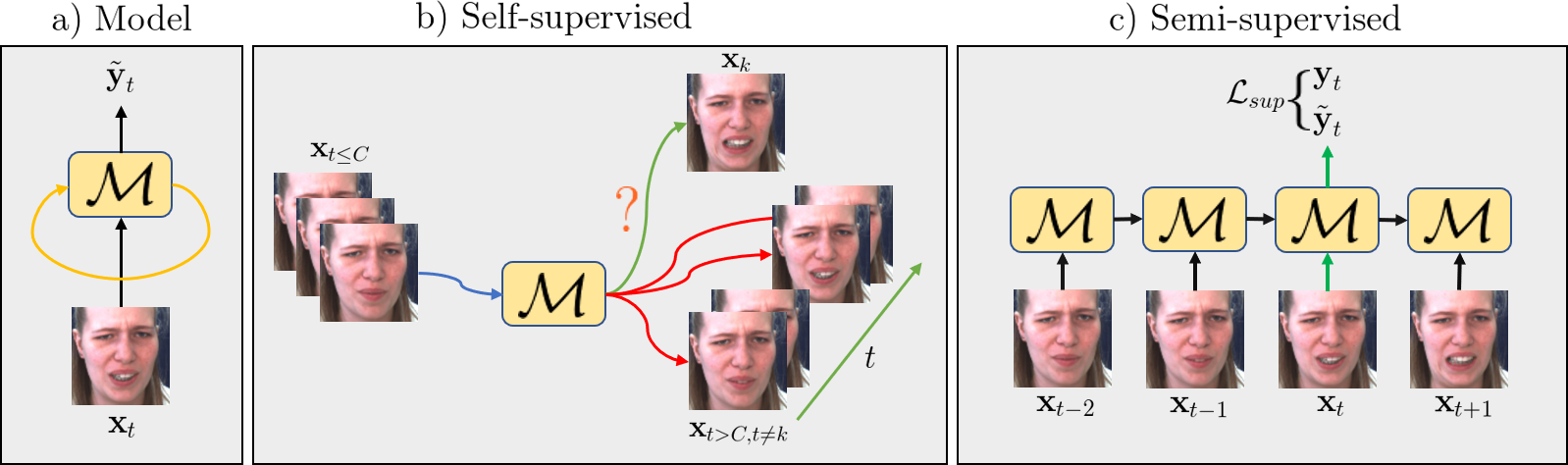}\par
  
  \caption{Overview of our proposed approach. a) We want to learn a temporal model for facial Action Unit intensity estimation that can be learned with a few \textit{randomly chosen} set of annotations. b) We propose to first learn a strong feature representation through \textit{contrastive learning}, by training the model to solve a proxy task, that requires no labeling. The proxy task uses a simple non-linear predictive function, and aims at predicting the feature representation that is similar to that of a specific target frame ${\bf x}_k$, and at the same time dissimilar to the rest of the given sequence. c) Once the model is trained to produce rich facial feature representations, we can further train it for facial Action Unit intensity estimation by using sequences in which the labeled frames are in random position in the sequence. }
  \label{graphabs}
\end{figure*}

\noindent While plenty of works exist for AU detection (i.e. whether an AU occurs or not in an image, regardless of its intensity), the more challenging task of automatically estimating their intensity has received less attention. Recent advances in supervised methods incorporate a variety of techniques including attention~\cite{sanchez2018,wu16,zhao2016}, co-occurrence modeling~\cite{walecki17,Zhang_2018_CVPR_BN}, or temporal dynamics~\cite{jaiswal2016,chu2017}. All these methods however require a large number of training instances to work properly, which entails the time consuming task of dataset labeling. This is even more profound for the problem of AU detection and intensity estimation where labeling is typically performed at a frame-level (i.e. each video frame must be labeled). Our goal in this paper is to devise a method that can effectively estimate AU intensity even when a very small (of the order of $2 \%$) and randomly chosen set of frames is used for model training.

This line of work has been pursued by the research community only recently ~\cite{li2019,Zhang_2018_CVPR_BN,Zhang_2018_CVPR,Zhang_2018_CVPR_weakly,Zhang_2019_CVPR,Zhang_2019_ICCV}. While these works have shown remarkable results, they still have some limitations: 1) They work on a per-frame basis, by learning a strong image-based feature representation that can later be used for AU intensity estimation. While some of these methods impose a temporal smoothing through ranking~\cite{Zhang_2019_CVPR,Zhang_2018_CVPR_weakly}, they still aim at learning a per-frame representation. 2) Some of these methods also require having a very specific set of annotations in hand~\cite{Zhang_2018_CVPR,Zhang_2018_CVPR_weakly,Zhang_2019_CVPR,Zhang_2019_ICCV}. In particular, they work by assuming that annotations are available for ``peak" and ``valley" frames, i.e. frames corresponding to a local maximum or minimum on the intensity. Identifying these frames requires a qualitative labeling of sequences before the annotation step. That is, while using peak and valley frames is effective with as little as $2\%$ of annotated frames, these annotations require indeed evaluating segments of videos which is also an expensive operation. 

In this paper, we take a different path in semi-supervised AU intensity estimation. We firstly assume a practical setting where only a very sparse (of the order of $2\%$) and randomly chosen set of frames (of a given dataset) is labeled. Our aim is to train a model which, even if trained on this sparse set of annotated frames, it can make dense per-frame predictions of AU intensity for all frames in a given test sequence. To this end, we build a model that combines a feature extraction network (ResNet-18,~\cite{he2016}) with a GRU unit~\cite{cho2014learning}, and a regressor head on top of the GRU that can make predictions for each frame and train it using back-propagation through time using only the predictions of the network at the sparsely labeled frames. We found however that training this model from scratch in an end-to-end manner using only a small number of labeled frames is a rather difficult task. Hence, we further propose to firstly train the backbone (i.e. the feature extractor and the GRU unit) in an unsupervised way on different unlabeled datasets, using the recent framework of contrastive learning~\cite{oord2018representation,ye2019unsupervised,wu2018unsupervised,he2019momentum,henaff2019data,tian2019contrastive,bachman2019learning}. The backbone is trained end-to-end with a contrastive loss on unlabeled videos of facial behavior, and used thereafter to train a model for AU intensity estimation using only few labels. An overview of our approach is shown in Fig.~\ref{graphabs}.

Our \textbf{main contributions} can be summarized as follows:
\begin{itemize}
\setlength\itemsep{0.1em}
\item We are the first to propose a temporal modeling of facial actions that can be learned with a sparse set of discontinuously and randomly chosen annotated facial images. Our practical approach to AU intensity estimation requires as little as $2\%$ of annotated data.

\item We are the first to apply the framework of contrastive learning for semi-supervised AU intensity estimation. We propose a two-stage pipeline where a model for obtaining a spatio-temporal facial representation is firstly trained on large unlabeled datasets of facial behavior using a contrastive learning formulation, and, then, the model is effectively trained for the task of AU intensity estimation using a small number of sparsely annotated video frames.

\item Our approach achieves state-of-the-art results on both BP4D and DISFA, when using a randomly chosen subset of $2\%$ frames. 
\end{itemize}

\section{Related work}

\subsection{Action Unit modeling}
The majority of existing works in facial Action Unit intensity estimation work on a fully supervised way, i.e. by assuming that a large amount of labeled data is available\cite{martinez2017automatic,ertugrul2020crossing}. Existing supervised methods are often split into methods that exploit the geometric structure of faces, also referred to as patch-based~\cite{zhao2016,li2017c,zhao2016facial}, methods that exploit the temporal correlation of AUs~\cite{jaiswal2016,chu2017}, and those that exploit the correlation that exists between different Action Units~\cite{walecki17,ertugrul2019,rudovic13b,rudovic13a,rudovic2015}. Other methods attempt to exploit different types of correlation~\cite{ming2015,li2017b,chu2019,yang2019,wang2019deep,eleftheriadis2016,wu16}.
Most recent methods reporting state of the art results on Action Unit intensity estimation build on AutoEncoders~\cite{tran2017}, or on Heatmap Regression~\cite{sanchez2018,ntinou2020transfer}. 

While there is a vast amount of literature in fully supervised methods for Action Unit intensity estimation, few works have focused on the more challenging task of addressing the same goal in a semi-supervised manner. In \cite{li2019}, a twin autoencoder is used to disentangle facial movements from head pose, the learned representation is then used for Action Unit detection. In \cite{Zhang_2018_CVPR}, an ordinal relevance regression method is applied. In \cite{Zhang_2018_CVPR_BN}, the relation between emotions and Action Units is used to generate a knowledge-graph that allows the use of the emotion labels to train the AU detector without labels. In \cite{wang_2019} a Restricted Boltzmann Machine and Support Vector Regression approach to model the AUs is proposed. In \cite{Zhang_2018_CVPR_weakly}, a knowledge-based approach is proposed, exploiting the temporal variation that exists between peak and valley frames. In \cite{Zhang_2019_CVPR}, the temporal ranking is exploited in a similar way, and a novel ADMM approach is used to enforce different constraints. In \cite{Zhang_2019_ICCV}, a learnable context matrix is used for each AU, which combined with several patch-based feature fusion is capable of learning AU intensity from key frames only. The majority of these methods however impose temporal constraints based on the fact that annotations are available for \textit{peak} and \textit{valley} frames, i.e. frames that correspond to a local maximum and minimum, respectively. Our method bypasses this need by applying a self-supervised pre-learning step, that learns a strong feature representation. Our approach uses as little as $2\%$ of annotated data. However, contrary to these methods, we use a \textit{randomly chosen} set of data. 

\subsection{Self-supervised learning in Computer Vision}
Self-supervised learning, often referred to as unsupervised learning, is an active research topic in Machine Learning. It involves defining a pretext or proxy task, with a corresponding loss, that leads to strong feature representation. This pretext task does not require the data to be labeled, e.g. it can be predicting relative location of image patches~\cite{doersch2015unsupervised}, predicting rotation from images~\cite{gidaris2018unsupervised}, sorting frames in a shuffled video~\cite{misra2016shuffle}, denoising~\cite{vincent2008extracting} or colorization~\cite{zhang2016colorful}, or pseudo-labeling through clustering~\cite{caron2018deep}. 

More recently, a number of state-of-the-art self-supervised methods based on the so-called contrastive learning formulation have been proposed ~\cite{oord2018representation,wu2018unsupervised,ye2019unsupervised,he2019momentum,henaff2019data,tian2019contrastive,han2019video,bachman2019learning}.
The idea is to define a loss for unsupervised learning which maximizes the similarity between the feature representations of two different instances of the same training sample while simultaneously minimizes the similarity with the representations computed from different samples. Among these methods, in this work, we build upon the Contrastive Predictive Coding of \cite{oord2018representation} which allows for a temporal model to be learned in an unsupervised manner from video data, and hence it is particular suitable for modeling facial behavior dynamics. To our knowledge, we are the first building upon this model for unsupervised modeling of facial behavior. We also show how to apply this model for semi-supervised AU intensity estimation.

\section{Method}
\label{sec:method}

We are interested in learning a spatio-temporal model for facial Action Unit intensity estimation, that is capable of modeling the temporal dynamics of expressions, as well as their feature representation, and that can be learned with few labels. Thus, we first introduce the problem statement and notation in Sec.~\ref{ssec:notation}, then we devote Sec.~\ref{ssec:network} to describing the model that will produce a structured output representation, and present our two-stage approach to the model learning in Sec.~\ref{ssec:cpc} and Sec.~\ref{ssec:train}.

\subsection{Problem statement and notation}
\label{ssec:notation}
We are interested in learning a model capable of predicting the intensity of some Action Units in a given sequence, that at the same time captures the spatial features responsible of displayed expressions and models the temporal correlation between them. We want such a model to be learned in a scenario where only a small set of frames are annotated with Action Unit intensity. 

Let ${\bf X} = \{\x_t\}_{t=1}^T$ be a sequence of $T$ video frames, where $\x_t \in \Re^{3 \times H \times W}$ is an RGB image of size $H \times W$. Our goal is to learn a model $\mathcal{M}$ that produces a structured output ${\bf \h{Y}} = \{\h{\y}_t\}_{t=1}^T$, where each $\h{\y}_t \in \Re^N$ corresponds to the predicted intensity of each of the $N$ Action Units of interest.  In a fully supervised setting, we would be given a set of ground-truth labels $\y_t$ for each sequence, and thus the model $\m{M}(\cdot \,; \theta)$ could be learned through regression and backpropagation through time~\cite{werbos1990backpropagation}. However, learning the model $\m{M}$ is a hard task, and requires a vast amount of per-frame labeled sequences. Also, training deep temporal models is often a challenging task, very sensitive to the initialization. We now describe our approach to learning the model efficiently with few labels. 

\subsection{Network}
\label{ssec:network}

Our model $\m{M}$ is split into a feature extraction block $f(\cdot \, ;\theta_f)$, a temporal module $g(\cdot \, ; \theta_g)$, and a final regressor head $c(\cdot\, ;\theta_c)$, with $\theta$ the parameters of each module. The choice for the \textit{feature extractor} $f(\cdot \,; \theta_f)$ is that of a ResNet-18~\cite{he2016}, without the last fully connected layer. \begin{wrapfigure}[17]{L}{3.8cm}
\vspace{-20pt}
  \begin{center}
    \includegraphics[width=0.92\textwidth]{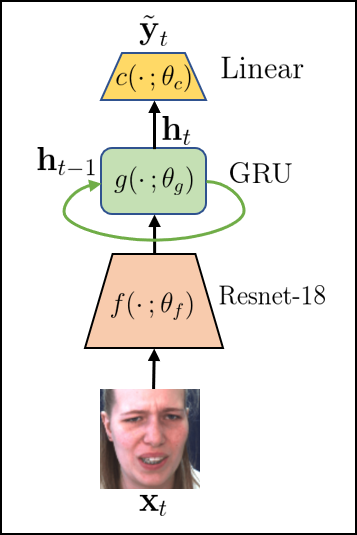}
  \end{center}
  \vspace{-5pt}
  \caption{Network: the model consists of a \textit{feature extractor}, a \textit{temporal block}, and a \textit{regressor head}. }
\end{wrapfigure}In order to keep a locally-based receptive field for the extracted features, we remove the average pooling layer. To reduce the complexity of the network, the last convolutional block is set to $256$ channels rather than the typical $512$. This way, for an input resolution of $128 \times 128$, the network produces a feature representation of $256$ channels, and of a spatial resolution of $8\times8$, i.e. $f(\cdot\, ; \theta_f) \in \Re^{256\times 8\times8}$. This way, each spatial output \textit{attends} a region of $16\times16$ pixels in the input image. 

The form of the \textit{temporal block} $g(\cdot\,; \theta_g)$ is that of a Convolutional Gated Recurrent Unit (GRU~\cite{cho2014learning}), with the latent dimension set to be the same as for the feature representation, i.e. $256\times8\times8$. The GRU receives, at time $t$, the input feature representation $f(\x_t; \theta_f) \in \Re^{256\times8\times8}$, along with the previous hidden state ${\bf h}_{t-1} \in \Re^{256\times8\times8}$, and propagates the hidden state at time $t$ as ${\bf h}_t = g( f( \x_t ; \theta_f), {\bf h}_{t-1}; \theta_g) \in \Re^{256\times8\times8}$. The GRU is modified to receive a tensor rather than a vector, by setting all the internal gates to be convolutional layers, with kernel size $1\times1$. 

Finally, a \textit{regressor head} $c(\cdot \, ; \theta_c)$ is placed on top of the GRU to perform a per-frame Action Unit intensity estimation. This head takes the hidden state at time ${\bf h}_t$, and after applying an average pooling operation, forwards the $256$-dimensional frame representation to a simple block consisting of a Batch Normalization layer~\cite{ioffe2015batch}, and a linear layer that produces an $N$-dimensional output, where $N$ is the number of the target AUs.

\subsection{Unsupervised pre-training}
\label{ssec:cpc}
It is worth noting that, even when working in a fully supervised manner, training the above model is a rather hard task. For the purposes of semi-supervised learning, we propose to use instead a self-supervised pre-training inspired by the contrastive learning~\cite{oord2018representation}, which allows us to make the network produce strong facial feature representations with no labels. 

Our learning goal is defined through a proxy \textit{predictive function} $p_k(\cdot \, ; \theta_p)$, tasked with predicting the future feature representations $f(\x_{t+k} ; \theta_f)$, at some time $t+k$ from a given contextual information $g( f(\x_t ; \theta_f) ; \theta_c)$, computed up to time $t$. The learning goal is to make $p_k\left(g\left(f(\x_t )\right)\right)$\footnote{We drop the dependency on the parameters $\theta$ for the sake of clarity} similar to $f(\x_{t+k})$, and at the same time different from other feature representations computed at the same time step $k$ for a different image $\x'$, $f(\x'_{t+k})$, at the same image $\x$ but at a different time step $k'$, $f(\x_{t+k'})$, and at a different image $\x'$ and time step $k'$, $f(\x'_{t+k'})$. With $p$ being a simple non-linear function, the learning burden lies on a feature representation capable of predicting the future. While in \cite{oord2018representation} the time step $k$ is fixed, in \cite{han2019video} the time $t+k$ is added recursively from $t+1$ to $t+k$. %

Put formally, for a sequence ${\bf X}=\{\x_t\}_{t=1}^T$, the first $C$ frames will represent the \textit{context}, whereas the last $P=T-C$ frames will be used for the \textit{predictive} task. Throughout our experiments, $T=15$ frames, and $C$ and $P$ are set to $10$ and $5$ frames, respectively, i.e. from the context estimated for the first $10$ frames, the goal is to predict the next $5$ feature representations. 

\begin{figure*}[t!]
    \centering\includegraphics[width=1\linewidth]{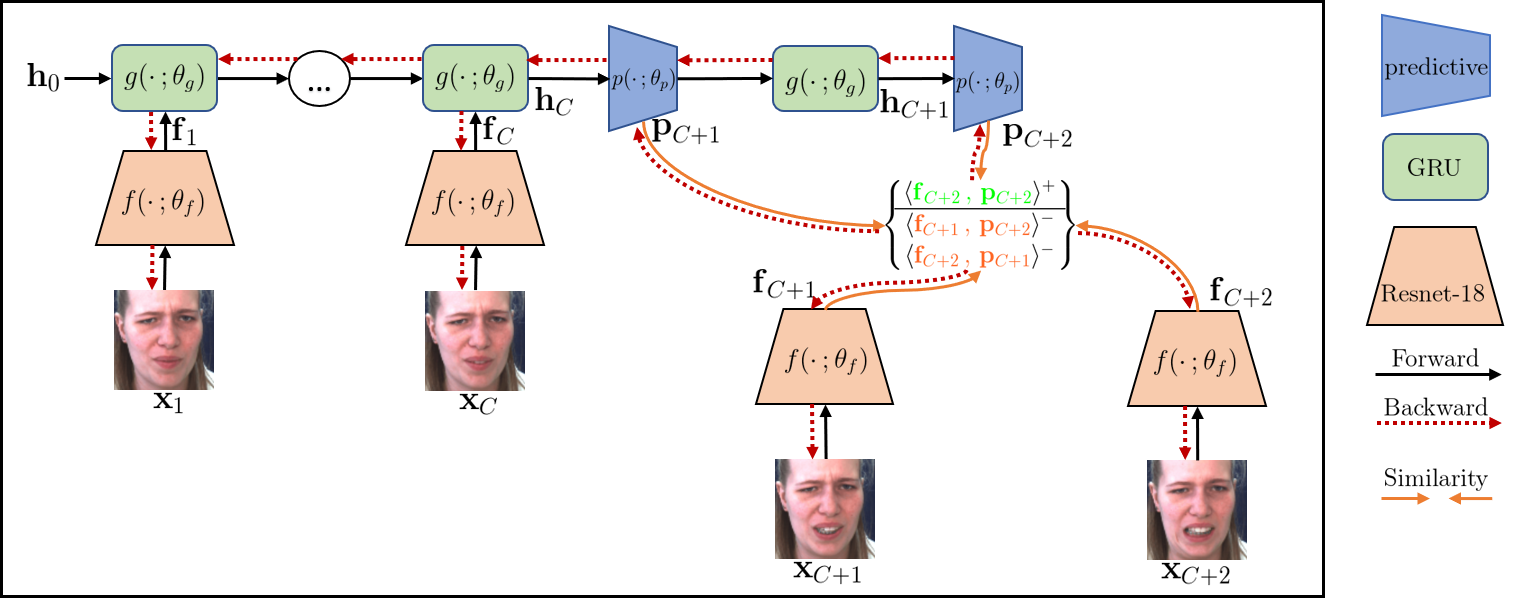}\par
  
  \caption{Self-supervised learning through predictive coding. A predictive function uses the hidden state ${\bf h}_C$ of the GRU, at time $t_C$, to predict the feature representation ${\bf p}_{C+1}$ at the next temporal step $t_{C+1}$. This prediction is also used to obtain the next hidden state ${\bf h}_{C+1}$. The feature prediction is then repeated to obtain ${\bf p}_{C+2}$. The pair given by ${\bf p}_{C+2}$ and ${\bf f}_{C+2}$ (in green), is used as a positive sample, whereas all other mixed pairs are used as negatives (orange pairs). For the sake of clarity we illustrate only one positive pair. All other similar correspondences act also as positives during training, see Sec~\ref{ssec:cpc}. We can then backpropagate (red lines) the similarity scores for the given pairs w.r.t. the contrastive loss defined in Eqn.~\ref{eq:cpc}, and learn a strong feature representation without labels. }
  \label{problem}
\end{figure*}

Let ${\bf f}^{(i,j)}_{t}$ represent the feature representation $f(\x_t)$ at time $t$, at the spatial location $(i,j)$. Similarly, let ${\bf p}^{(i,j)}_{t+k}$ represent the output of the predictive function $p_k(g(f(\x_t)))$, at time $t+k$, at the spatial location $(i,j)$. We use a recursive context generation: 
\begin{eqnarray}
    {\bf p}_{t+1} &=& p_1(g(f(\x_t)))\nonumber\\
    \dots & & \dots \dots \nonumber\\
    {\bf p}_{t+k} &=& p_1(g({\bf p}_{t+k-1}))\nonumber
\end{eqnarray}
i.e. we enforce the predictions to be conditioned not only to on previous observations, but also on the previous predictions. Recall that $g(f(\x_t)) = g(f(\x_t), {\bf h}_{t-1})$, i.e. all frames before $t$ are summarized in the context at time $t$. 

The learning is accomplished through a Noise Contrastive Estimation~\cite{hadsell2006dimensionality,gutmann2010noise}, where the goal is to classify real from noisy samples. Real samples are in practice formed by pairs $({\bf f}^{(i,j)}_{t+k}, {\bf p}^{(i,j)}_{t+k})$, while noisy samples are formed by pairs $({\bf f}^{(i',j')}_{t+k'}, {\bf p}^{(i,j)}_{t+k})$. In other words, noisy samples are formed by pairs composed of the feature representation at time $t$ and spatial location $(i,j)$, and all the predictions taken from the same time position at different spatial locations, the predictions taken at different time steps, and even the predictions computed at different images in a given batch. For a given set of $P$ predicted representations with size $H \times W$, the loss is formulated as:
\begin{equation}
\label{eq:cpc}
    \mathcal{L}_{nce} = - \sum_{k,i,j} \left[ \log \frac{e^{\langle{\bf f}^{(i,j)}_{t+k}, {\bf p}^{(i,j)}_{t+k}\rangle}}{\sum_{k',i',j'} e^{\langle{\bf f}^{(i',j')}_{t+k'}, {\bf p}^{(i,j)}_{t+k}\rangle}} \right]
\end{equation}
where $\langle \cdot, \cdot \rangle$ denotes the dot product, and is used as a similarity score between the feature representations. In Eqn.~\ref{eq:cpc}, $k \in \{1 \dots P\}$, and $(i,j) \in \{(1,1) \dots (H,W)\}$. The loss within the brackets represents the typical cross-entropy objective used for classification, where the goal is to classify the positive pair among a set of $P \times H \times W$ classes (i.e. pairwise scores). When the set of negatives is enhanced with other images in the batch, the set of possible classes becomes $B \times P \times H \times W$, with $B$ the batch size. In our setting, $B=20$, $P=5$, and $W=H=8$, as aforementioned, making the number of predictions be $6400$. Note that in \cite{oord2018representation} the number of classes is dozens. This makes our predictive task much harder, making the representations be more locally distinct. The reason behind this approach relies on that we want the representation to be strong for a downstream task based on a \textit{per-frame} classification. Minimizing Eqn.~\ref{eq:cpc} leads to learning the weights $\theta_f$ and $\theta_g$ without requiring labeled videos. This procedure allows us to learn a strong representation with videos collected in-the-wild. A visual description is shown in Fig.~\ref{problem}. 

\subsection{Learning with partially labeled data}
\label{ssec:train}
\begin{figure*}[t!]
    \centering\includegraphics[width=1\linewidth]{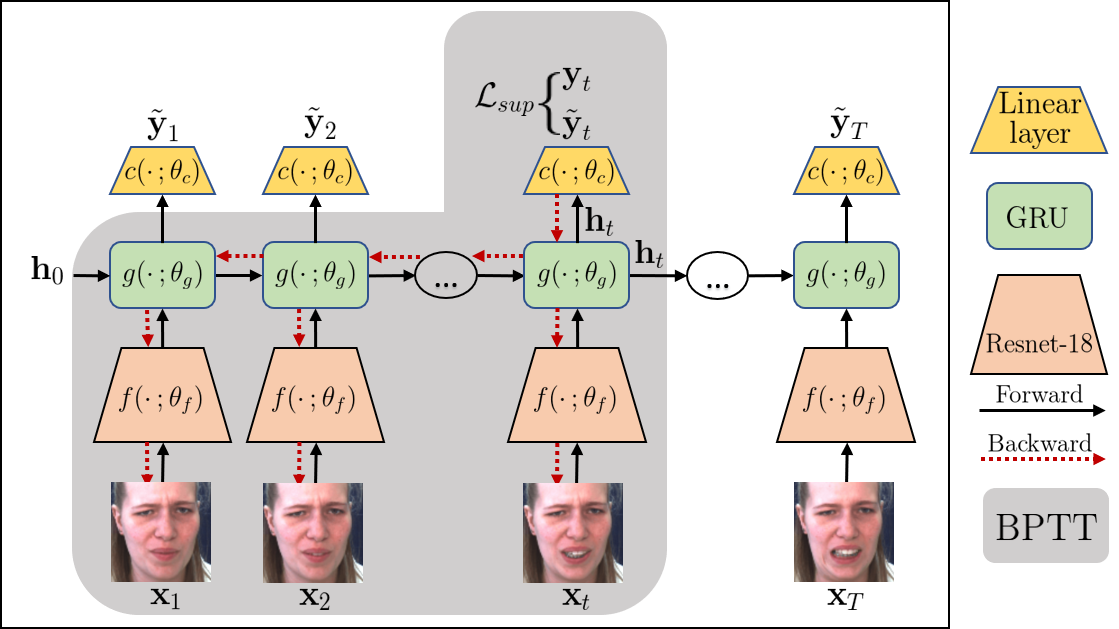}\par
  
  \caption{Learning with partially labeled data: the network produces a structured output $\{\tilde{\y}_t\}_{t=1}^T$, but only one frame $\y_t$ is labeled. We apply Backpropagation through time (BPTT) w.r.t. the labeled frame. While the labeled frames remain the same through the training process, their position in a queried sequence is randomly shifted (i.e. the number of frames before and after the labeled frame are randomly varied).}
  \label{learn}
\end{figure*}

 Now, we turn into how to train the network when only a few labeled facial images, randomly and discontinuously sampled, are available. In this paper, we propose a simple approach that consists of choosing random windows around a labeled frame, so that the position of the latter in the sequence varies each time it is queried for updating the network parameters. Using this approach, we apply back-propagation w.r.t. the output at the labeled frame. In other words, if only the frame $\x_t$ is labeled within a sequence ${\bf X} = \{\x\}_{t=1}^T$, the classifier $c(\cdot \, ; \theta_c)$ is updated only with a given labeled frame, and the feature extractor and temporal units, $f(\cdot \, ;\theta_f)$ and $g(\cdot \, ; \theta_g)$, are updated through back-propagation through time up to the labeled frame $t$. Denoting with $\y_t$ the label for frame $t$, the loss is formulated as:
 \begin{equation}
     \mathcal{L}_{sup} = \| \y_t - c(g(f(\x_t))) \|^2,
 \end{equation}
 where it is important to recall the dependency of $c(g(f(\x_t)))$ with all frames up to $t$. This way, the parameters $\theta_f$ and $\theta_g$ are updated with all the frames in the sequence up to $t$, and the parameters $\theta_c$ are updated only with the context given by $g$ at time $t$ and the corresponding labeled frame. This approach is illustrated in Fig.~\ref{learn}.

\section{Experimental results}
\noindent \textbf{Data} Throughout our experiments, we use three different datasets. For the self-supervised pretraining described in Section~\ref{ssec:cpc}, we used the raw videos from the \textbf{Aff-Wild2} database~\cite{kollias2017recognition,kollias2018aff,kollias2018multi,kollias2019deep,kollias2019expression,kollias2020analysing,zafeiriou2017aff} (i.e. with no annotations). We use $422$ videos from Aff-Wild2 with around $1.2$ million frames in total. In order to learn the feature representation using the predictive task, we used 351 videos for training, and 71 videos for validation. For the Action Unit intensity estimation, we use the \textbf{BP4D} and \textbf{DISFA} datasets. The \textbf{BP4D} database~\cite{zhang2014} is the main corpus of the \textbf{FERA2015}~\cite{valstar2015}, and consists of videos of $41$ subjects performing $8$ different tasks, making it a total of $328$ videos. For our experiments, we use the official train and validation subject-independent partitions, consisting of $21$ and $20$ subjects, respectively. The database contains around $140,000$ frames, and is annotated with AU intensity for $5$ AUs. In addition to the BP4D, we evaluate our method and perform a thorough ablation study on the \textbf{DISFA} database~\cite{mavadati2013}, which includes $27$ videos, each of a different subject, while performing computer tasks, for an average of $\sim4$ minutes. It comprises around $130,000$ frames annotated with the intensity of $12$ AUs. To compare our method w.r.t. state-of-the-art results, we perform a subject independent three-fold cross-validation, where, for each fold, $18$ subjects are used for training and $9$ for testing. For our ablation studies, we use only one of the three folds (the same for all studies). \\

\begin{figure*}[h!]
    \centering\includegraphics[width=0.45\linewidth]{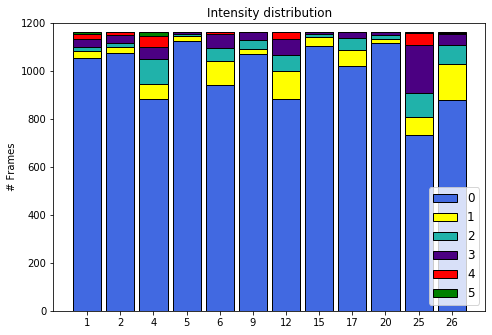} 
    \centering\includegraphics[width=0.45\linewidth]{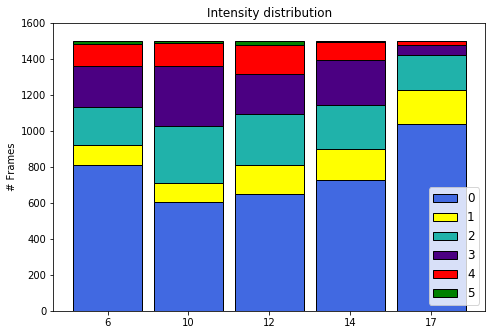} 
  
  \caption{AU intensity distribution for DISFA (left) and BP4D (right).}
  \label{distribution}
\end{figure*}

\noindent \textbf{Frame selection} Following existing works on semi-supervised Action Unit intensity estimation, we consider a labeled set of $2\%$ of frames. However, contrary to existing works, we select the labeled frames \textit{randomly}. To avoid the results of our ablation studies to be different due to the choice of data, we use the same subset of images to train both our models and those used for ablation studies. The number of training images for BP4D is $1498$, whereas the number of images for each fold in DISFA is $1162$. Fig.~\ref{distribution} shows the AU intensity distribution. \\

\noindent \textbf{Evaluation metrics} We use the Intra-Class Correlation (ICC(3,1)~\cite{shrout79}), commonly used to rank methods in existing benchmarks. For an AU $j$ with ground-truth labels $\{y^j_i\}_{i=1}^N$, and predictions $\{\tilde{y}^j_i\}_{i=1}^N$, the ICC score is defined as $ICC^j = \frac{W^j-S^j}{W^j+S^j}$, with $W^j = \frac{1}{N}\sum_{i} \left((y^j_i - \hat{y}^j)^2 + (\tilde{y}^j_i - \hat{y}^j)^2 \right)$, $S^j = \sum_{i} (y^j_i - \tilde{y}^j_i)^2$, and $\hat{y}^j = \frac{1}{2N}\sum_i (y^j_i + \tilde{y}^j_i)$. In addition, we report the Mean Absolute Error (MAE). \\

\noindent \textbf{Set up} We use the PyTorch automatic differentiation package in all our experiments~\cite{paszke2017}. We use the publicly available S$^3$FD face detector of ~\cite{Zhang_2017_ICCV} to detect a facial bounding box in each video. The images are then tightly cropped around the center and resized to $128\times128$. Some augmentation is applied to the images for both the self-supervised training and the final stage (see below). \\

\noindent \textbf{Self-supervised training} The training is composed of $33509$ sequences of $15$ frames each, sampled with a stride of 2, and extracted from Aff-Wild2. To avoid discontinuities, sequences where a face is not detected are discarded. The number of valid test sequences to validate the performance of the predictor is $7164$. To minimize non-facial content we tightly crop around the center of the bounding box. We apply a set of heavy augmentations during training, including uniform rotation ($\pm 20$ deg.), uniform scaling ($\pm 10\%$), random flipping, random jitter, and random hue, contrast, saturation and brightness. The number of context frames is set to $10$, and the set of predicted frames is $5$. We use the Adam optimizer~\cite{kingma2014}, with an initial learning rate of $10^{-3}$ and weight decay of $10^{-5}$. The model is trained for $300$ epochs using 8 GPUs, each having a batch size of $20$ sequences. The training takes approximately 1 day to be completed. We evaluate the capacity of the network to pick up the right pair by measuring the Top-$n$ accuracy on the validation set, with $n=1,3,5$. In other words, Top-$n$ shows the percentage of samples in the validation set where the score corresponding to the positive pair was among the highest $n$ scores. The results are shown in the second row of Table~\ref{tab:comparison}. As an ablation study, we also used a subset of DISFA to perform the self-supervised training. See Section~\ref{ssec:ablation} for further details. The Top-$n$ accuracy for the DISFA is shown in the first row of Table~\ref{tab:comparison}. \\

\begin{wraptable}[7]{r}{5cm}
\centering
\vspace{-12pt}
\begin{tabular}{|l|c|c|c|}
         \hline
         \rowcolor{Gray}Accuracy & \multicolumn{1}{|c|}{Top-1} & \multicolumn{1}{|c|}{Top-3} & \multicolumn{1}{|c|}{Top-5} \\
         \hline \hline
         DISFA & .165 & .467 & .695 \\
         Aff-Wild2 & .303 & .651& .781 \\
         \hline
    \end{tabular}
\caption{Accuracy on the validation set of DISFA and Aff-Wild2 for the Contrastive Predictive task. See Sec~\ref{ssec:ablation} for details}
\label{tab:comparison} 
\end{wraptable}

\noindent \textbf{Semi-supervised training} To train the Action Unit intensity regressor, we use learning rate $10^{-5}$ with weight decay $10^{-5}$. We use Adam with $\beta=(0.5,0.999)$. For the model trained from scratch, we use a Kaiming initialization~\cite{He_2015_ICCV}. We observed that the model is sensitive to the initialization. When using normally distributed random weights, the training was unstable, and the performance was very poor.  

\subsection{Ablation study}
\label{ssec:ablation}
We first perform an ablation study to validate the effectiveness of our proposed approach w.r.t. different alternatives. To this end, we use the most challenging fold from DISFA to train and validate each of the models, i.e. we used 18 subjects to train and 9 to test. The best performing model for each method is chosen. In particular, we illustrate the results of the following options in Table~\ref{tab:au_results2} and Table~\ref{tab:au_results3}. 
\begin{itemize}
    \item \textbf{R18 - $2\%$} We train a simple ResNet-18 on the same $2\%$ of the data, by minimizing the $\mathcal{L}_2$ loss between the available labels and the predictions on the corresponding images. We use the same augmentation as mentioned above.
    \item \textbf{R18 - Sup.} We train the same ResNet-18 on the full dataset ($58140$ images). 
    \item \textbf{R18+GRU-scratch - $2\%$} We train the whole pipeline without the self-supervised training described in Section~\ref{ssec:cpc}, i.e. we train the whole model from scratch. 
    \item \textbf{R18+GRU-scratch Sup.} We also trained the whole pipeline without the self-supervised training in a fully supervised manner.%
    \item \textbf{R18+GRU-scratch Pseudo-GT} In this setting, we used a different approach to semi-supervised learning, that of pseudo-labeling. In particular, we used the R18 - $2\%$ model to generate the labels for the training set, and used these labels to train our model. In addition, the weights of the R18-$2\%$ network are also used to initialize the R18+GRU pipeline. 
    \item \textbf{Ours(*DISFA) - $2\%$} In this setting, we used the same training partition to perform the self-supervised training described in Section~\ref{ssec:cpc}, i.e. we used the $18$ training videos from DISFA. Then, we initialized our network with the generated weights, and trained using only the aforementioned $2\%$ of labels. 
    \item \textbf{Ours - $2\%$} This setting corresponds to using our proposed approach to learn the model with $2\%$ of the data.
    \item \textbf{Ours - Sup.} We finally evaluate the performance of the model when the whole training set is available. This serves as an upper bound in the performance of the proposed approach. 
\end{itemize} 

\noindent \textbf{Discussion} The results of all these models are shown in Table~\ref{tab:au_results2} and Table~\ref{tab:au_results3}. From these results, we can make the following observations:
\begin{itemize}
    \item \textit{1. Accuracy of self-supervised learning} The results shown in Table~\ref{tab:comparison} indicate that the performance in the predictive task is superior when the network is trained on Aff-Wild2 than when it is trained on DISFA. We attribute this difference to the fact that DISFA has less variability than Aff-Wild2, mainly due to the recording conditions. In addition, the training set for DISFA is composed of barely $18$ videos, and hence the number of negative samples are highly likely to include segments that are too similar. In this scenario, the negative samples become almost indistinguishable from the positive ones, making the predictive task harder to learn. 
    \item \textit{2. Importance of pre-text training set} In addition to the lower performance on the predictive task, it is worth noting that the learned representation when using few videos is much weaker than that learned from videos collected in the wild. This is illustrated in the performance that the network yields in the task of facial Action Unit intensity estimation. It is important to remark that collecting videos in-the-wild that \textit{do not} require labeling is nowadays cheaper than annotating all the frames with $12$ Action Units even for a small number of videos. 
    \item \textit{3. Influence of self-supervised learning} We observe that, whether on DISFA or on Aff-Wild2, the results on the downstream task are remarkably better than training the network from scratch. We observe this not only for our case of interest where few labels are available, but also when training the network in a fully supervised manner. We also observed that, when trained from scratch, the network is certainly sensitive to the initialization. This has been observed in \cite{He_2015_ICCV}, that indicates that a poor initialization with deep networks can lead to vanishing gradients, making the training process unstable. In our experiments, we observe that training the Resnet-18 in a supervised manner without any temporal modeling yields better results than training the whole pipeline, also in a supervised manner. We can see that our method is effective both when training with few labels and when training in a supervised way. 
    \item \textit{4. Influence of pseudo-labeling} We observe that this technique can be powerful enough, yielding an ICC score similar to that given by training the network from scratch in a fully supervised manner. 
\end{itemize}

\begin{table*}[t!]
\begin{center}
\begingroup
\setlength{\tabcolsep}{3.5pt}
\resizebox{\textwidth}{!}{%
\begin{tabular}{|p{0.001\textwidth}>{\raggedleft} l|*{12}{c}|c|}
  \hline
  \rowcolor{Gray} & Dataset & \multicolumn{13}{c|}{DISFA - Only $2\%$ of frames are annotated}\\
  \rowcolor{Gray}
		& AU & \multicolumn{1}{c}{1} & \multicolumn{1}{c}{2} & \multicolumn{1}{c}{4} & \multicolumn{1}{c}{5} & \multicolumn{1}{c}{6} & \multicolumn{1}{c}{9} & \multicolumn{1}{c}{12} & \multicolumn{1}{c}{15} & \multicolumn{1}{c}{17} & \multicolumn{1}{c}{20} & \multicolumn{1}{c}{25} & \multicolumn{1}{c|}{26} & Avg.\\
	\hline \hline
	\multirow{4}{*}{\rotatebox{90}{\noindent \small{ICC}}} 
    & R18 $2\%$ & \bbold{.144} & -.014 & .375 & \bbold{.129} & .499 & \bbold{.366} & .737 & .208 & .355 & \bbold{.210} & .901 & .458 & .364 \\
    & R18+GRU-scratch $2\%$  & -.032 & -.126 & .340 & .009 & .393 & .043 & .650 & .077 & .048 & .004 & .841 & .434 & .223\\
    & Ours($^*$Disfa)-$2\%$ &.084 & \bbold{.070} & .489 & .033 & .481 & .225 & \bbold{.772} & \bbold{.337} & .304 & .127 & .853 & \bbold{.578} & .363\\
    & Ours ($^*$Affwild2)-$2\%$  & -.010 & -.022 & \bbold{.657} & .068 & \bbold{.566} & .358 & .737 & .291 & \bbold{.366} & .109 & \bbold{.944} & .537 & \bbold{.383}\\
    \hline 
    \multirow{4}{*}{\rotatebox{90}{\noindent \small{MAE}}}
    & R18 $2\%$ & \bbold{.209} & \bbold{.228} & 1.022 & \bbold{.042} & .339 & .311 & .333 & \bbold{.137} & .306 & .191 & .307 & .474 & .325 \\
    & R18+GRU-scratch $2\%$  & .555 & .512 & .984 & .076 & .442 & .397 & .542 & .178 & .252 & \bbold{.150} & .546 & .436 & .423\\
    & Ours($^*$Disfa)-$2\%$  & .238 & .236 & \bbold{.683} & .098 & .329 & .287 & \bbold{.326} & .174 & \bbold{.305} & .217 & .428 & .389 & \bbold{.309} \\
    & Ours($^*$Affwild2)-$2\%$  &.392 & .463 & .754 & .129 & \bbold{.304} & \bbold{.277} & .360 & .189 & .386 & .201 & \bbold{.260} & \bbold{.298} & .334\\
    \hline
\end{tabular}
}
\endgroup
\end{center}
\caption{Ablation study on DISFA in a scenario where only $2\%$ of frames are annotated. R18-2\% refers to a simple ResNet-18 trained with 2\% of the data. R18+GRU-scratch-2\% refers to the method described in Section~\ref{ssec:train}, without the self-supervised learning. Ours(*Disfa) refers to our method, with the self-supervised learning stage being done on the DISFA dataset. Ours(*Affwild2)-2\% refers to our method, trained on $2\%$ of the data.\bbold{bold} indicates best performance}
\label{tab:au_results2}
\end{table*}

\begin{table*}[t!]
\begin{center}
\begingroup
\setlength{\tabcolsep}{3.5pt}
\resizebox{\textwidth}{!}{%
\begin{tabular}{|p{0.001\textwidth}>{\raggedleft} l|*{12}{c}|c|}
  \hline
  \rowcolor{Gray} & Dataset & \multicolumn{13}{c|}{DISFA - All frames are annotated}\\
  \rowcolor{Gray}
		& AU & \multicolumn{1}{c}{1} & \multicolumn{1}{c}{2} & \multicolumn{1}{c}{4} & \multicolumn{1}{c}{5} & \multicolumn{1}{c}{6} & \multicolumn{1}{c}{9} & \multicolumn{1}{c}{12} & \multicolumn{1}{c}{15} & \multicolumn{1}{c}{17} & \multicolumn{1}{c}{20} & \multicolumn{1}{c}{25} & \multicolumn{1}{c|}{26} & Avg.\\
	\hline \hline
	\multirow{4}{*}{\rotatebox{90}{\noindent \small{ICC}}} 
    & R18 Sup. & .215 & .030 & .470 & \bbold{.339} & \bbold{.583} & .382 & .790 & \bbold{.387} & .476 & .383 & .887 & .565 & .459 \\
    & R18+GRU-scratch Sup. & .047 & -.056 & .366 & .097 & .455 & .145 & .775 & .089 & .313 & .126 & .812 & .501 & .306 \\ 
    & R18+GRU-Pseudo-GT  & .096 & .017 & .469 & .115 & .470 & .227 & .653 & .186 & .317 & .257 & .578 & .250 & .303 \\ 
    & Ours ($^*$Affwild2)-Sup.  & \bbold{.291} & \bbold{.107} & \bbold{.495} & .281 & .499 & \bbold{.424} & \bbold{.798} & .327 & \bbold{.513} & \bbold{.473} & \bbold{.908} & \bbold{.648} & \bbold{.480} \\
    \hline
    \multirow{4}{*}{\rotatebox{90}{\noindent \small{MAE}}}
    & R18 Sup. & .216 & .222 & 1.002 & .048 & \bbold{.302} & .302 & .314 & \bbold{.117} & \bbold{.200} & .134 & .321 & \bbold{.346} & .294 \\
    & R18+GRU-scratch Sup. & .365 & .341 & 1.022 & .113 & .364 & .553 & .352 & .204 & .444 & .259 & .601 & .549 & .431\\
    & R18+GRU-Pseudo-GT  & .225 & .374 & .816 & .062 & .640 & .282 & .530 & .296 & .844 & .185 & 1.127 & .631 & .501 \\
    & Ours ($^*$Affwild2)-Sup.  & \bbold{.161} & \bbold{.200} & \bbold{.815} & \bbold{.043} & .334 & \bbold{.273} & \bbold{.292} & .127 & .215 & \bbold{.108} & \bbold{.291} & .362 & \bbold{.268}\\
	\hline
\end{tabular}
}
\endgroup
\end{center}
\caption{Ablation study on DISFA where a fully supervised setting is applied. R18 Sup. corresponds to a Resnet-18 trained in a supervised manner. R18+GRU-scratch-Sup. refers to the supervised training. R18+GRU-scratch-PseudoGT refers to a fully supervised training using the pseudo-labels produced by the R18-2\% shown in Table~\ref{tab:au_results2}. Ours(*Affwild2)-Sup. refers to our method, trained with all the labels. \bbold{bold} indicates best performance. }
\label{tab:au_results3}
\end{table*}

\subsection{Comparison with state-of-the-art}
We now show the results of our method w.r.t. reported state of the art methods on weakly supervised learning for facial Action Unit intensity estimation. We report our results for both the BP4D case and for the three-fold cross validation performed in DISFA, so as to make our results comparable to existing works. We show the results of our method for BP4D in Table~\ref{tab:resultsfera15test}, and the results for DISFA in Table~\ref{tab:au_disfa}. 
We compare our method with the most recent works reporting Action Unit intensity estimation with partially labeled data. In particular, we compare the performance of our method with that of \textbf{KBSS}~\cite{Zhang_2018_CVPR_weakly}, \textbf{KJRE}~\cite{Zhang_2019_CVPR}, and \textbf{CFLF}~\cite{Zhang_2019_ICCV}. Importantly, the three methods require the labeling of key frames, mainly due to the fact that one of the components for weakly supervised learning relies on assuming that intermediate frames between keys follow some monotonic behavior. Also, both the KBSS and CFLF work under the basis of a $1\%$ of labeled frames. However, these methods, reportedly, use a different percentage of frames \textit{per Action Unit}. Different from these methods, our proposed approach works with a randomly chosen subset of data. As shown in both Table~\ref{tab:resultsfera15test} and Table~\ref{tab:au_disfa}, our method yields competitive results, and surpasses the three aforementioned methods in terms of average ICC. 

\begin{table}[htbp]
\begin{center}
\begingroup
\setlength{\tabcolsep}{7pt}
\resizebox{0.8\textwidth}{!}{%
\begin{tabular}{|p{0.001\textwidth}>{\raggedleft} l|*{5}{c}|c|}
  \hline
  \rowcolor{Gray} & Dataset & \multicolumn{6}{c|}{BP4D}\\
  \rowcolor{Gray}
		& AU & \multicolumn{1}{c}{6} & \multicolumn{1}{c}{10} & \multicolumn{1}{c}{12} & \multicolumn{1}{c}{14} & \multicolumn{1}{c|}{17} & \multicolumn{1}{|c|}{Avg.}\\
	\hline \hline
	 	\multirow{4}{*}{\rotatebox{90}{\noindent ICC}} %
	& KBSS~\cite{Zhang_2018_CVPR_weakly}* & .760 & .725 & .840 & .445 & .454 & .645 \\
	& KJRE~\cite{Zhang_2019_CVPR}* $6\%$& .710 & .610 & \bbold{.870} & .390 & .420 & .600 \\
    & CFLF~\cite{Zhang_2019_ICCV}* & \bbold{.766} & .703 & .827 & .411 & \bbold{.600} & .661 \\
    & Ours $2\%$ & \bbold{.766} & \bbold{.749} & .857 & \bbold{.475} & .553 & \bbold{.680} \\
    \hline 
    \multirow{4}{*}{\rotatebox{90}{\noindent MAE}}
    & KBSS~\cite{Zhang_2018_CVPR_weakly}* & .738 & \bbold{.773} & .694 & .990 & .895 & .818 \\
    & KJRE~\cite{Zhang_2019_CVPR}* $6\%$ & .820 & .950 & \bbold{.640} & 1.080 & .850 & .870 \\
    & CFLF~\cite{Zhang_2019_ICCV}* & \bbold{.624} & .830 & .694 & 1.000 & \bbold{.626} & \bbold{.741} \\
    & Ours $2\%$ & .645 & .913 & .826 & \bbold{.979} & .628 & .798\\
     \hline
\end{tabular}
}
\endgroup
\end{center}
\caption{Intensity estimation results on BP4D. (*) Indicates results taken from reference. \bbold{bold} indicates best performance. }
\label{tab:resultsfera15test}
\end{table}

\begin{table*}[htp]
\label{au_disfa}
\begin{center}
\begingroup
\setlength{\tabcolsep}{3.5pt}
\resizebox{\textwidth}{!}{%
\begin{tabular}{|p{0.001\textwidth}>{\raggedleft} l|*{12}{c}|c|}
  \hline
  \rowcolor{Gray} & Dataset & \multicolumn{13}{c|}{DISFA}\\
  \rowcolor{Gray}
		& AU & \multicolumn{1}{c}{1} & \multicolumn{1}{c}{2} & \multicolumn{1}{c}{4} & \multicolumn{1}{c}{5} & \multicolumn{1}{c}{6} & \multicolumn{1}{c}{9} & \multicolumn{1}{c}{12} & \multicolumn{1}{c}{15} & \multicolumn{1}{c}{17} & \multicolumn{1}{c}{20} & \multicolumn{1}{c}{25} & \multicolumn{1}{c|}{26} & Avg.\\
	\hline \hline
	\multirow{4}{*}{\rotatebox{90}{\noindent \small{ICC}}} 
    & KBSS~\cite{Zhang_2018_CVPR_weakly}* & .136 & .116 & .480 & .169 & .433 & .353 & .710 & .154 & .248 & .085 & .778 & .536 & .350 \\
    & KJRE~\cite{Zhang_2019_CVPR}* $6\%$ & .270 & \bbold{.350} & .250 & .330 & .510 & .310 & .670 & .140 & .170 & .200 & .740 & .250 & .350 \\
	& CLFL~\cite{Zhang_2019_ICCV}* &  .263 & .194 & .459 &  \bbold{.354} & .516 & \bbold{.356} & .707 & \bbold{.183} & \bbold{.340} & \bbold{.206} & .811 & .510 & .408 \\ 
	& Ours $2\%$ & \bbold{.327} & .328 & \bbold{.645} & -.024 & \bbold{.601}
        & .335 &  \bbold{.783}  & .181 & .243 & .078 &
         \bbold{.882} &  \bbold{.578} & \bbold{.413} \\
    \hline 
    \multirow{4}{*}{\rotatebox{90}{\noindent \small{MAE}}}
    &  KBSS~\cite{Zhang_2018_CVPR_weakly}* & .532 & .489 & .818 & .237 & .389 & .375 & .434 & .321 & .497 & .355 & .613 & .440 & .458 \\
	& KJRE~\cite{Zhang_2019_CVPR}* $6\%$ & 1.020 &.920 &1.860 &.700 &.790 &.870 &.770 &.600 &.800 &.720 &.960 &.940 & .910 \\
	& CLFL~\cite{Zhang_2019_ICCV}* &  \bbold{.326} & \bbold{.280} & \bbold{.605} & \bbold{.126} & \bbold{.350} & \bbold{.275} & \bbold{.425} & \bbold{.180} & \bbold{.290} & \bbold{.164} & .530 & .398 & \bbold{.329} \\ 
	& Ours $2\%$ & .430 & .358 & .653 & .194 & .381 & .370 & .457 & .247 & .376 & .212 &
        \bbold{.446} & \bbold{.387} & .376 \\
	\hline
\end{tabular}
}
\endgroup
\end{center}
\caption{Intensity estimation results on DISFA. (*) Indicates results taken from reference. \bbold{bold} indicates best performance. }
\label{tab:au_disfa}
\end{table*}

\section{Conclusion}
\label{sec:conclusion}
In this paper, we proposed a novel approach to semi-supervised training of facial Action Unit intensity estimation, that is capable of delivering competing results with as little as $2\%$ of annotated frames. To this end, we proposed a self-supervised learning approach that can capture strong semantic representations of the face, that can later be used to train models in a semi-supervised way. Our approach surpasses existing works on semi-supervised learning of Action Unit intensity estimation. We also demonstrated that our approach, when used in a fully supervised manner, largely outperforms a model trained from scratch, thus demonstrating that our approach is also valid for supervised learning. The experimental evaluation proved the effectiveness of our method, through several ablation studies. 

%
%
\bibliographystyle{splncs04}
\bibliography{accv2020submission}
\end{document}